\documentclass[11pt]{article}

\usepackage[final]{acl}

\usepackage{times}
\usepackage{latexsym}

\usepackage[T1]{fontenc}
\usepackage[utf8]{inputenc}

\usepackage{microtype}

\usepackage{inconsolata}

\usepackage{graphicx}

\usepackage{subfigure}
\usepackage{bbm}
\usepackage{amssymb}
\usepackage{amsmath}
\usepackage{booktabs}
\usepackage{color, colortbl}
\usepackage{multirow}
\usepackage{listings}

\usepackage{algorithm}
\usepackage{algpseudocode}
\usepackage{tabularx}

\newcolumntype{Y}{>{\centering\arraybackslash}X}

\definecolor{lightcyan}{rgb}{0.88, 1, 1}
\definecolor{rankcolor}{HTML}{1b65cc}
\title{MoRA: High-Rank Updating for Parameter-Efficient Fine-Tuning}

\author{
\textbf{Ting Jiang}\textsuperscript{\rm 1} , \textbf{Shaohan Huang}\textsuperscript{\rm 2} , \textbf{Shengyue Luo} \textsuperscript{\rm 2}, \textbf{Zihan Zhang}\textsuperscript{\rm 2} , \textbf{Haizhen Huang}\textsuperscript{\rm 2}\\
\textbf{Furu Wei}\textsuperscript{\rm 2} \textbf{,} \textbf{Weiwei Deng}\textsuperscript{\rm 2} \textbf{,} \textbf{Feng Sun}\textsuperscript{\rm 2} \textbf{,} \textbf{Qi Zhang}\textsuperscript{\rm 2} \textbf{,} \textbf{Deqing Wang}\textsuperscript{\rm 1}$^\dagger$ \textbf{,} \textbf{Fuzhen Zhuang}\textsuperscript{\rm 1} \\
\textsuperscript{\rm 1}Beihang University \textsuperscript{\rm 2}Microsoft Corporation\\
\texttt{royokong@buaa.edu.cn}\\
}

\begin{document}
\maketitle
\begin{abstract}

Low-rank adaptation (LoRA) is a popular parameter-efficient fine-tuning (PEFT) method for large language models (LLMs).
%In this paper, we analyze the influence of low-rank updating in LoRA and find it limits the capability of LLMs to learn new knowledge or memory.
In this paper, we analyze the impact of low-rank updating, as implemented in LoRA. Our findings suggest that the low-rank updating mechanism may limit the ability of LLMs to effectively learn and memorize new knowledge.
Inspired by this observation, we propose a new method called MoRA, which employs a square matrix to achieve high-rank updating while maintaining the same number of trainable parameters.
To achieve it,
we introduce the corresponding non-parameter operators to reduce the input dimension and increase the output dimension for the square matrix.
Furthermore, these operators ensure that the weight can be merged back into LLMs, which makes our method can be deployed like LoRA.
We perform a comprehensive evaluation of our method across five tasks: instruction tuning, mathematical reasoning, continual pretraining, memory and pretraining. Our method outperforms LoRA on memory-intensive tasks and achieves comparable performance on other tasks.
Our code will be available at \url{https://github.com/kongds/MoRA}.
%In addition, we also compare our method with ReLoA on pretraining to demonstrate the effectiveness of our method.
\end{abstract}

\section{Introduction}

As the size of language models increases, parameter-efficient fine-tuning (PEFT)~\cite{houlsby2019parameter} has emerged as a popular technique to adapt these models to specific downstream tasks.
Compared to Full Fine-Tuning (FFT), which updates all model parameters, PEFT modifies only a small part of the parameters. For example, it can achieve similar performance with FFT by updating less than 1\% of the parameters in some tasks~\cite{hu2021lora}, which significantly reduces the memory requirements for the optimizer and facilitates the storage and deployment of fine-tuned models.

%Among the existing PEFT methods, Low-Rank Adaptation (LoRA) is a prevalent method especially for Large Language Models (LLM). By updating the parameters via low rank matrices, LoRA achieves better performance compared to other PEFT methods such as prompt tuning~\cite{lester2021power} or adapters~\cite{houlsby2019parameter}. Moreover, the low rank matrices can be merged back into the original model parameters, which does not introduce additional computation cost during inference.
%Compared to FFT, LoRA achieves even better performance on text classification tasks like GLUE~\cite{wang2018glue}.
%And most successor methods of LoRA mainly verify their efficient based on GLUE by achieving better performance or less trainable parameters.
%To better evaluate the performances for LLM, recent methods~\cite{liu2024dora, meng2024periodiclora, zhu2024asymmetry} leverage instruction datasets such as Alpaca~\cite{wang2024far} or reasoning tasks like GSM8K~\cite{cobbe2021training}.

%However, even many efforts to improve LoRA, LoRA still the frist choice to fine-tuning LLM with PEFT benfit from simple and efficient.
%Due to different settings and datasets, it makes hard to understand the progression of these methods.
%For example, there are methods applying LoRA to all linear layers in base model, while some applying to query and value linear in self attention.

\begin{figure}[t]
    \centering

    \subfigure[LoRA (\textcolor{rankcolor}{$r = 8$})]
    {
        \includegraphics[width=0.4385\columnwidth]{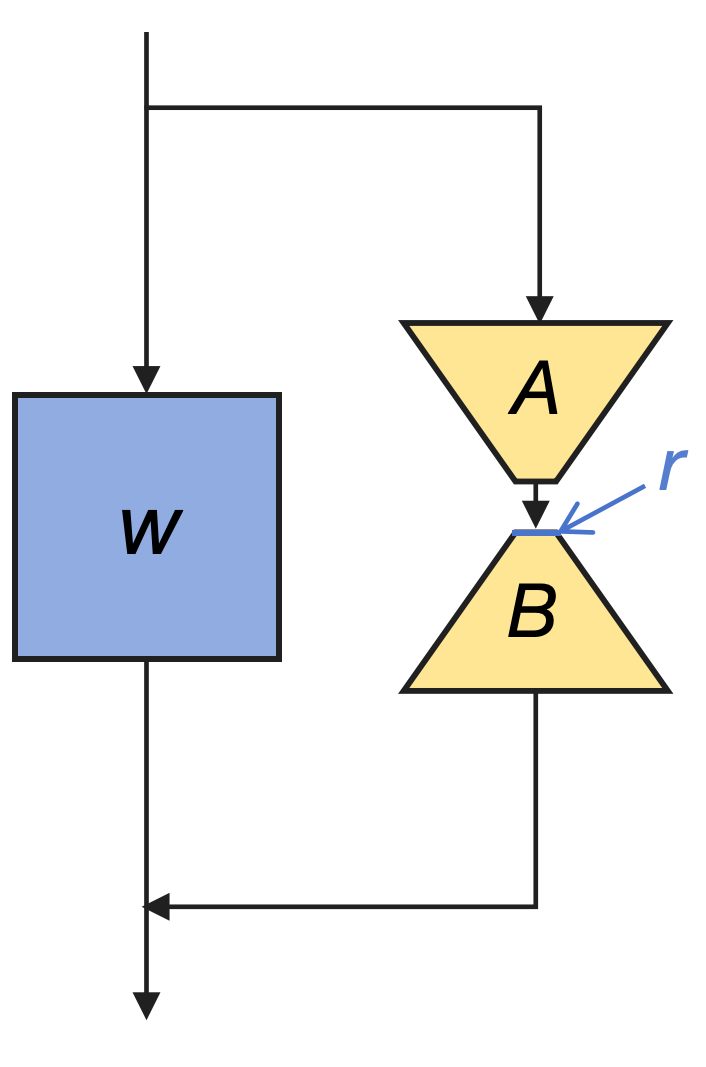}
    }
    \subfigure[MoRA (\textcolor{rankcolor}{$r = 256$})]
    {
        \includegraphics[width=0.4615\columnwidth]{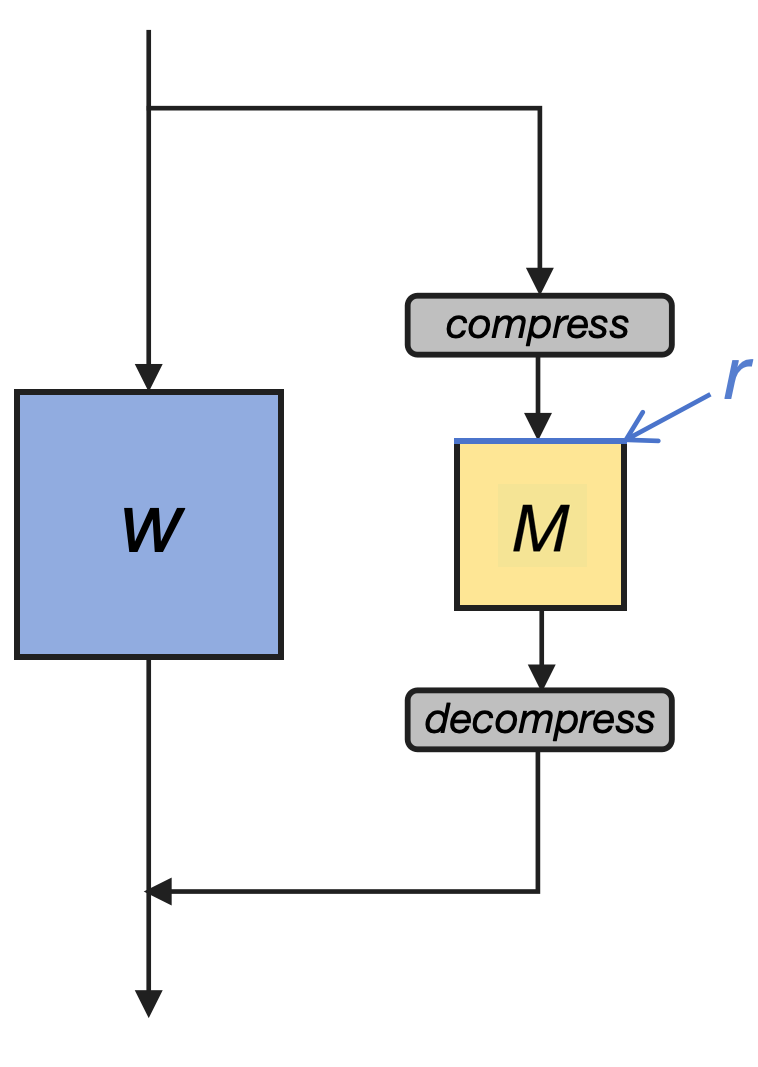}
    }
    \caption{
      An overview of our method compared to LoRA under \textbf{same} number of trainable parameters.
    $W$ is the frozen weight from model. $A$ and $B$ are trainable low-rank matrices in LoRA.
      $M$ is the trainable matrix in our method.
      Gray parts are non-parameter operators to reducing the input dimension and increasing the output dimension.
      \textcolor{rankcolor}{r} represents the rank in two methods.
      }
    \label{fig:framework}
\end{figure}

Among the existing PEFT methods, Low-Rank Adaptation (LoRA)~\cite{hu2021lora} is particularly prevalent for LLMs. LoRA enhances performance over other PEFT methods such as prompt tuning~\cite{lester2021power} or adapters~\cite{houlsby2019parameter} by updating parameters via low-rank matrices. These matrices can be merged into the original model parameters, thereby avoiding additional computational costs during inference. %When compared to FFT, LoRA demonstrates superior performance on text classification tasks like GLUE~\cite{wang2018glue}.
There are numerous methods that aim to improve LoRA for LLMs. However,
most methods primarily validate their efficiency based on GLUE~\cite{wang2018glue}, either by achieving better performance or by requiring fewer trainable parameters.
Recent methods~\cite{liu2024dora, meng2024periodiclora, zhu2024asymmetry} leverage instruction tuning task such as Alpaca~\cite{wang2024far} or reasoning tasks like GSM8K~\cite{cobbe2021training} to better evaluate their performance on LLMs.
However, the diverse settings and datasets used in the evaluation complicate the understanding of their progression.
%For instance, some methods~\cite{shi2024reslora} apply LoRA on query and value linears in self-attention module, while others~\cite{liu2024dora} apply LoRA to all linear layers.
%Moreover, certain methods~\cite{kopiczko2023vera, renduchintala2023tied} strive to reduce the number of trainable parameters even further by freezing or sharing low-rank matrices. However, this reduction in trainable parameters even slow down the training time compared to the original LoRA method~\cite{kopiczko2023vera}, making it less applicable.

% 783 744

% In this paper, we conduct a comprehensive evaluation of LoRA in various tasks under same settings, including instruction tuning, mathematical reasoning, and continual pretraining. We find LoRA like methods show similar performance on these tasks and observe they perform comparably to FFT in instruction tuning but falls short in mathematical reasoning and continual pretraining.
% %Despite this, instruction tuning has been widely used to evaluate the efficiency of LoRA variants in LLMs~\cite{meng2024periodiclora, liu2024dora, kopiczko2023vera}.
% Among these tasks, instruction tuning primarily focuses on interacting with the format, rather than on acquiring knowledge and capabilities, which are learned almost entirely during pretraining~\cite{zhou2024lima}.
% We find LoRA is easy to learn to follow response formats in instruction tuning, but struggles with other tasks that require enhancing knowledge and capabilities through fine-tuning.

In this paper, we conduct a comprehensive evaluation of LoRA across various tasks under the same settings, including instruction tuning, mathematical reasoning, and continual pretraining. We find that LoRA-like methods demonstrate similar performance across these tasks and they perform comparably to FFT in instruction tuning but fall short in mathematical reasoning and continual pretraining. Among these tasks, instruction tuning primarily focuses on interacting with the format, rather than acquiring knowledge and capabilities, which are learned almost entirely during pretraining~\cite{zhou2024lima}. We observe that LoRA is easily adapted to follow response formats in instruction tuning but struggles with other tasks that require enhancing knowledge and capabilities through fine-tuning.

One plausible explanation for this limitation observed with LoRA could be its reliance on low-rank updates~\cite{lialin2023stack}. The low-rank update matrix, $\Delta W$, struggles to estimate the full-rank updates in FFT, particularly in memory-intensive tasks like continual pretraining that require memorizing domain-specific knowledge.
Since the rank of $\Delta W$ is significantly smaller than the full rank, this limitation restricts capacity to store new information via fine-tuning. Moreover, current variants of LoRA cannot alter the inherent characteristic of low-rank updates.
%Although current methods such as ReLoRA~\cite{lialin2023stack} and COLA~\cite{xia2024chain} attempt to augment the rank by merging LoRA into LLMs during training,
%they still use low-rank updates to estimate the full rank.
To validate this, we conducted a memorization task using pseudo-data to assess the performance of LoRA in memorizing new knowledge. We found that LoRA performed significantly worse than FFT, even with a large rank such as 256.
 %As shown in Figure~\ref{fig:mtloss}, LoRA struggles to memorize new knowledge compared to FFT, even with a large rank like 256.

Given these observations, we introduce a method called MoRA, which employs a square matrix as opposed to low-rank matrices, aiming to maximize the rank in $\Delta W$ while maintaining the same number of trainable parameters. For instance, when utilizing 8 rank with the hidden size 4096, LoRA employs two low-rank matrices $A\in \mathbb{R}^{4096 \times 8}$ and $B\in \mathbb{R}^{8 \times 4096}$, with $rank(\Delta W) \leq 8$. Under same number of parameters, our method uses a square matrix $M \in \mathbb{R}^{256 \times 256}$, with $rank(\Delta W) \leq 256$, as depicted in Figure~\ref{fig:framework}.
Notably, our method exhibits a greater capacity than LoRA with a large rank.
To decrease the input dimension and increase the output dimension for $M$, we develop corresponding non-parameter operators.
Furthermore, these operators and $M$ can be substituted by a $\Delta W$, ensuring our method can be merged back into LLM like LoRA.
%To validate the effectiveness of our method, we conduct a evaluation across five tasks: instruction tuning, mathematical reasoning, continual pretraining, and memory. Our method shows better performance on memory-intensive tasks and comparable performance on other tasks compared to LoRA.

Our contributions are as follows:

\begin{enumerate}
\item We introduce MoRA, a novel method that employs a square matrix instead of low-rank matrices in LoRA to achieve high-rank updating, while maintaining the same number of trainable parameters.
\item  We discuss four kinds of non-parameter operators of MoRA to reduce the input dimension and increase the output dimension for the square matrix, while ensures that the weight can be merged back into LLMs.
\item
  We evaluate MoRA across five tasks: memory, instruction tuning, mathematical reasoning, continual pretraining, and pretraining. Our method outperforms LoRA on memory-intensive tasks and achieves comparable performance on other tasks, which demonstrates the effectiveness of high-rank updating.
\end{enumerate}

\section{Related Work}
\subsection{LoRA}
LoRA is one of the most popular PEFT methods for fine-tuning LLM, owing to its broad applicability and robust performance in comparison to other methods.
To approximate the updated weight $\Delta W$ in FFT, LoRA employs two low-rank matrices for its decomposition. By adjusting the rank of these two matrices, LoRA can accordingly modify the trainable parameters. Benefit from it, LoRA can merge these matrices after fine-tuning without the inference latency compared to FFT.
There are many methods to further improve LoRA, particularly for the application in LLMs. %Several works focus on enhancing performance.
DoRA\cite{liu2024dora} further decomposes the original weight into magnitude and direction components and uses LoRA to update the direction component. LoRA+\cite{Hayou2024} employs different learning rates for the two low-rank matrices to improve learning efficiency. ReLoRA\cite{lialin2023stack} integrates LoRA into the LLM during training to increase the rank of the final $\Delta W$.
%Another set of works aims to reduce trainable parameters while maintaining performance. \cite{kopiczko2023vera} freezes the low-rank matrices and trains the diagonal matrices added between low-rank matrices. \cite{renduchintala2023tied} ties the low-rank matrices across all layers of LLMs.

\subsection{Fine-Tuning with LLMs}
\label{sec:2.2}
Despite the impressive performance of LLMs with in-context learning, certain scenarios still necessitate fine-tuning, which can be broadly categorized into three types.
The first type, instruction tuning, aims to better align LLMs with end tasks and user preferences, without significantly enhancing the knowledge and capabilities of LLMs~\cite{zhou2024lima}. This approach simplifies the process of dealing with varied tasks and understanding complex instructions.
The second type involves complex reasoning tasks such as mathematical problem-solving~\cite{collins2023evaluating, imani2023mathprompter, yu2023metamath}, where general instruction tuning often falls short in handling complex, symbolic, multi-step reasoning tasks. To improve the reasoning abilities of LLMs, the majority of research focuses on creating corresponding training datasets, either by leveraging larger teacher models like GPT-4~\cite{fu2023specializing}, or by rephrasing questions along a reasoning path~\cite{yu2023metamath}.
The third type, continual pretraining~\cite{cheng2023adapting, chen2023disc, han2023medalpaca, liu2023chipnemo}, aims to enhance the domain-specific capabilities of LLMs. Unlike instruction tuning, it necessitates fine-tuning to augment the corresponding domain-specific knowledge and capabilities.

However, most variants of LoRA~\cite{kopiczko2023vera, lialin2023stack, dettmers2024qlora, zhu2024asymmetry} predominantly employ instruction tuning or text classification tasks from GLUE~\cite{wang2018glue} to validate their efficacy on LLMs. Given that instruction tuning requires the least capacity for fine-tuning compared to other types, it may not accurately reflect the effectiveness of LoRA variants.
To better evaluate their methods, recent works~\cite{meng2024periodiclora, liu2024dora, shi2024reslora, renduchintala2023tied} have employed reasoning tasks to test their methods.
But the training sets used are often too small for LLMs to effectively learn reasoning. For instance, some methods~\cite{meng2024periodiclora, renduchintala2023tied} utilize the GSM8K~\cite{cobbe2021training} with only 7.5K training samples. Compare to the SOTA method with 395K training  samples~\cite{yu2023metamath}, this small training set achieves worse performance on reasoning and makes it hard to evaluate the effectiveness of these methods.
%In term of performance, the latter achieves 66.5 accuracy on the GSM8K dataset compared to 28.0 for the former.

\section{Analysis the Influence of Low-rank Updating}
\label{sec:3}

%\subsection{LoRA and Low-rank Updating}
The key idea of LoRA~\cite{hu2021lora} involves the use of low-rank updates to estimate full-rank updates in FFT. Formally, given a pretrained parameter matrix $ W_0 \in \mathbb{R}^{d \times k} $, LoRA employs two low-rank matrices to calculate the weight update $\Delta W$:
\begin{equation}
h = W_0x + \Delta W x= W_0x + BAx
\end{equation}
where $A \in \mathbb{R}^{r \times k}$ and $B \in \mathbb{R}^{d \times r}$ represent the low-rank matrices in LoRA. To ensure that $\Delta W=0$ at the beginning of training, LoRA initializes $A$ with a Gaussian distribution and $B$ with zero. Due to the low-rank decomposition of $\Delta W$ into $BA$, the $rank(\Delta W) \leq r$.
%where $r \ll \min (k, r)$ to achieve the parameter-efficient fine-tuning.
The weight update in LoRA exhibits a markedly low rank, $r \ll \min (d, k)$, in comparison to the full-rank updating in FFT.
Low-rank updating by LoRA shows on-par performance with full-rank updating in some tasks such as text classification or instruction tuning~\cite{liu2024dora, meng2024periodiclora}.
However, for tasks like complex reasoning or continual pretraining, LoRA tends to show worse performance~\cite{liu2023chipnemo}.

%Based on above observations, we propose a hypothesis about low-rank updating for pertrained language models.
Based on these observations, we propose a hypothesis that low-rank updating is easy to leverage original knowledge and capabilities of LLM to solve task, but it is struggle to handle tasks that require enhancing knowledge and capabilities of LLM.

 \begin{figure}[t]
     \centering
         \includegraphics[width=0.99\columnwidth]{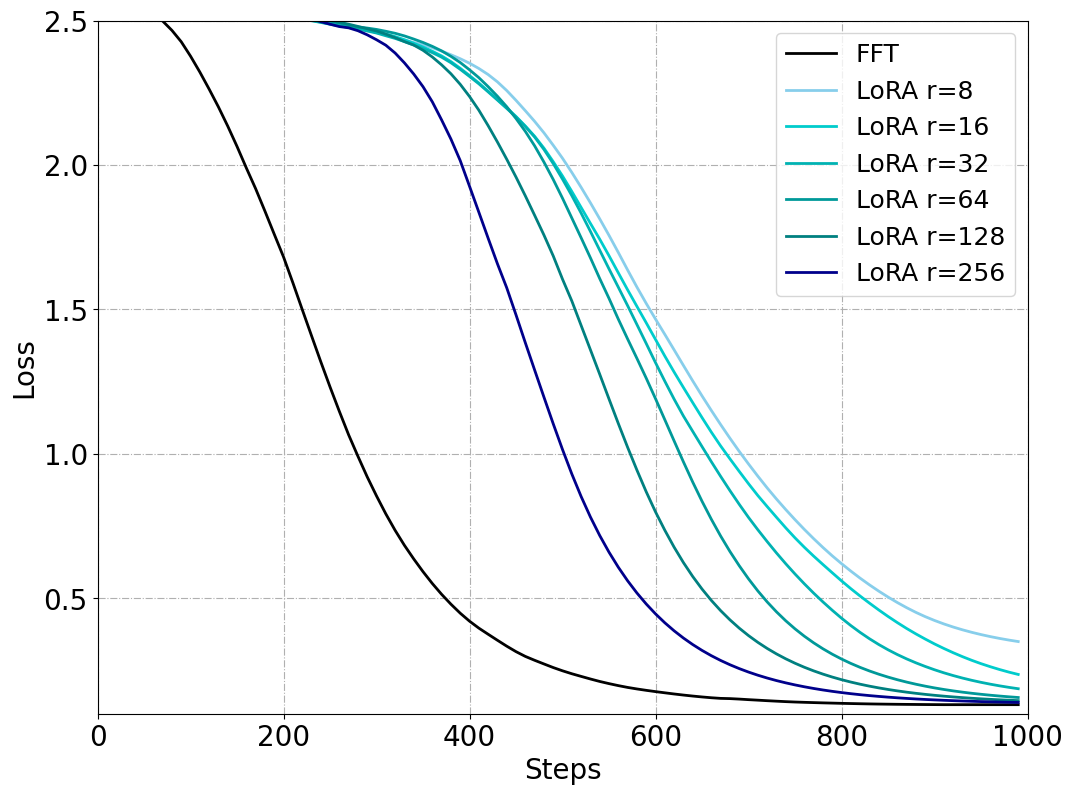}
     \caption{
       Performance of memorizing UUID pairs through fine-tuning with FFT and LoRA.
       }
     \label{fig:mtloss}
%\vspace{-10pt}
 \end{figure}
%\subsection{LoRA and Low-rank Updating}
To substantiate this hypothesis, we examine the differences between LoRA and FFT in terms of memorizing new knowledge through fine-tuning. In order to circumvent leveraging the original knowledge of the LLM, we randomly generate 10K pairs of Universally Unique Identifiers (UUIDs), each pair comprising two UUIDs with 32 hexadecimal values. The task requires the LLM to generate the corresponding UUID based on the input UUID. For instance, given a UUID such as ``205f3777-52b6-4270-9f67-c5125867d358'', the model should generate the corresponding UUID based on 10K training pairs.
This task can also be viewed as a question-answering task, while the knowledge indispensable for accomplishing it is exclusively from the training datasets rather than the LLM itself.

For the training settings, we employ LLaMA-2 7B as base model, utilizing 1,000 pairs per batch and conducting 100 epochs. %We use 3e-5 learning rate for FFT and 3e-4 for LoRA.
For the LoRA, we apply low-rank matrices to all linear layers and search learning rate from $\{$1e-4$,$2e-4$,$3e-4$\}$ to enhance performances.
We conduct the experiment on LoRA using various ranks $r \in \{8, 16, 32, 64, 128, 256\}$.
For the FFT, we directly use a learning rate of 3e-5.
Based on Figure~\ref{fig:mtloss}, we observe low-rank updating are hard to memorizing new knowledge compared to FFT.
Although constantly increasing the rank of LoRA can alleviate this problem, the gap still exists.

In contrast to the memory task, we also evaluate the performance gap between LoRA and FFT on instruction tuning, which merely introduces new knowledge.
Similar to previous results~\cite{meng2024periodiclora, zhu2024asymmetry}, we also find that LoRA matches the performance of FFT with small rank $r=8$ in Table~\ref{tab:main_results}.
This indicates that LoRA can easily leverage the original knowledge of LLMs by fine-tuning like FFT.

\section{Method}
\label{sec:41}

Based on the above analysis, we propose a new method to alleviate the negative effects of low-rank updating. The main idea of our method is to utilize the same trainable parameters as much as possible to achieve a higher rank in $\Delta W$.
Consider to the pretrained weight $W_0 \in \mathbb{R}^{d \times k}$, LoRA uses two low-rank matrices $A$ and $B$ with $(d+k)r$ total trainable parameters for rank $r$.
Under same trainable parameters, a square matrix $M \in \mathbb{R}^{\hat{r} \times \hat{r}}$ where $\hat{r} = \lfloor \sqrt{(d+k)r} \rfloor$ can achieve the highest rank due to $r \ll \min (d, k)$.

To accomplish this, we need to reduce the input dimension and increase the output dimension for $M$. Formally,
\begin{equation}
\label{eq:hle-bert}
\begin{split}
  %\hat{x} &= f_{\text{comp}}(x)\\
  %h &= W_0x + f_{\text{decomp}}(M\hat{x})
  h &= W_0x + f_{\text{decomp}}\left(M  f_{\text{comp}}\left(x\right)\right)
\end{split}
\end{equation}
where $f_{\text{comp}}: \mathbb{R}^k \rightarrow \mathbb{R}^{\hat{r}}$ denotes the function that decreases the input dimension of $x$ from $k$ to $\hat{r}$, and $f_{\text{decomp}}: \mathbb{R}^{\hat{r}} \rightarrow \mathbb{R}^{d}$ represents the function that enhances the output dimension from $\hat{r}$ to $d$.
Furthermore, these two functions ought to be non-parameterized operators and expected to execute in linear time corresponding to the dimension. %$\mathcal{O}(k)$ and $\mathcal{O}(d)$ floating-point operations (FLOPs) respectively.
They should also have corresponding function, $f_{\overline{\text{comp}}}: \mathbb{R}^{\hat{r}\times \hat{r}} \rightarrow \mathbb{R}^{\hat{r}\times k}$ and $f_{\overline{\text{decomp}}}: \mathbb{R}^{\hat{r}\times k} \rightarrow \mathbb{R}^{d \times k}$, to transform $M$ into $\Delta W$. For any $x$, the following should hold:
\begin{equation}
  \label{eq:3}
f_{\text{decomp}}\left(Mf_{\text{comp}}\left(x\right)\right) =  \Delta W x , \forall x \in \mathbb{R}^k \\
\end{equation}
where $ \Delta W = f_{\overline{\text{decomp}}}\left(f_{\overline{\text{comp}}}\left(M\right)\right)$.
If Eq.~\ref{eq:3} holds, $M$ can be losslessly expanded to $\Delta W$ based on $f_{\text{comp}}$ and $f_{\text{decomp}}$. This allows our method to merge back into the LLM like LoRA.
%We present an illustration in Figure~\ref{fig:mm_example}, where $M$ can be losslessly expanded to $\Delta W$ based on  $f_{\text{comp}}$ and $f_{\text{decomp}}$ functions, thereby ensuring that Eq.~\ref{eq:decomp} is satisfied.

% \begin{figure}[t]
%     \centering
%         \includegraphics[width=0.99\columnwidth]{mm.png}
%     \caption{
%       Performance of memorizing new knowledge through fine-tuning with FFT and LoRA.
%       }
%     \label{fig:mm_example}
% \end{figure}

%\subsection{Compress and Decompresss}

%Drawing on the paradigm depicted in Figure~\ref{fig:mm_example}, we delve into the design of these functions to enhance the performance.
For the design of $f_{\text{comp}}$ and $f_{\text{comp}}$, we explore several methods to implement these functions.
One straightforward method is truncating the dimension and subsequently add it in corresponding dimension.
Formally, this can be represented as:
\begin{equation}
\label{eq:cut}
\begin{split}
f_{\text{comp}}\left(x\right) &= x_{1:\hat{r}}\\
%f_{\text{decomp}}\left(x\right) &= \begin{bmatrix} x \\ \mathbf{0} \end{bmatrix}
f_{\text{decomp}}\left(x\right) &= \begin{bmatrix} x \\ \mathbf{0} \end{bmatrix}
\end{split}
\end{equation}
%where \( \mathbf{0}_{d-r} \) is a zero vector of length \( d-r \).
and the corresponding $\Delta W$ is:
\begin{equation}
\Delta W = \begin{bmatrix}
M & \mathbf{0} \\
\mathbf{0} & \mathbf{0}
\end{bmatrix}
\end{equation}
However, this method leads to a significant loss of information during compression and only modifies a segment of the output by appending a zero vector during decompression.
To improve it, we can share the rows and columns of $M$ to achieve a more efficient compression and decompression. Formally, this can be represented as:
\begin{equation}
  \label{eq:share}
\begin{split}
  %f_\text{comp}\left(x\right) &= \left[ \sum_{j \in g_i} x_j \right]_{i=1}^r\\
  f_\text{comp}\left(x\right) &= \begin{bmatrix} \sum_{j \in g_i} x_j \end{bmatrix}_{i=1}^r\\
  %f_\text{comp}\left(x\right) &= \left[ \Sigma_{j \in g_i} x_j \right]_{i=1}^r\\
  f_{\text{decomp}}\left(x\right) &= \begin{bmatrix} x_{\widetilde{g}'_i} \end{bmatrix}_{i=1}^d
\end{split}
\end{equation}
Here, $g$ and $g'$ represent predefined groups that share the same row and column in $M$, respectively. The $j\in g_i$ indicates that the $j$-th dimension belongs to the $i$-th group in $g$. The term $\widetilde{g}'_i$ is the reverse of $g'_i$, referring to the $i$-th dimension associated with the $\widetilde{g}'_i$-th group in $g'$. The corresponding $\Delta W$ is as follows:

\begin{equation}
\Delta W_{i,j} = M_{\widetilde{g}'_i, \widetilde{g}_j}
\end{equation}
Sharing rows and columns can be efficient for larger ranks such as $r=128$ or $r=256$, as only a few rows or columns in $\Delta W$ share a common row or column.
For instance, considering to  $\Delta W \in \mathbb{R}^{4096 \times 4096}$ for $r=128$, which has $\hat{r}=1024$ and $M \in \mathbb{R}^{1024 \times 1024}$. In this situation, only 4 rows or columns share the same row or column.
Conversely, for smaller ranks such as $r=8$, where $\hat{r}=256$, it requires average 16 rows or columns in a group to share the same row or column in $M$.
It can lead to inefficiencies due to the significant information loss during compression in Eq.~\ref{eq:share}.

To enhance performance for smaller ranks, we reshape $x$ instead of directly compressing it, to preserve the input information. In this context, $f_\text{comp}\left(x\right) : \mathbb{R}^k \rightarrow \mathbb{R}^{n \times \hat{r}}$ and  $f_{\text{decomp}}: \mathbb{R}^{n \times \hat{r}} \rightarrow \mathbb{R}^{d}$.
Corresponding $f_\text{comp}$, $f_{\text{decomp}}$ and $\Delta W$ are as follows:
\begin{equation}
  \small
  \label{eq:decaq}
\begin{split}
f_\text{comp}\left(x\right) &= \begin{bmatrix} x_{1:\hat{r}} & x_{\hat{r}: 2\hat{r}} & \cdots  &x_{(n-1)\hat{r}: n\hat{r}}  \end{bmatrix}\\
f_{\text{decomp}}\left(x\right) &= \text{concat}(x)\\
\Delta W &= \begin{bmatrix}
M & \mathbf{0} & \cdots & \mathbf{0} \\
\mathbf{0} & M & \cdots & \mathbf{0} \\
\vdots & \vdots & \ddots & \vdots \\
\mathbf{0} & \mathbf{0} & \cdots & M
\end{bmatrix}
\end{split}
\end{equation}
where $\text{concat}(x)$ refers to concatenate the rows of $x$ into a vector.
For simplicity, we omit the padding and truncation operators in above functions and focus on the case where $d=k$. %For the case of $d \neq k$, we can use same method with truncation and padding operators.
 In comparison to sharing columns and rows, this method incurs additional computational overhead by reshaping $x$ into $\mathbb{R}^{n \times \hat{r}}$ instead of $\mathbb{R}^{\hat{r}}$. However, given that the size of $M$ is significantly smaller than $W_0$, this additional computation is very small for rank like 8. For instance, when fine-tuning the 7B model with rank of 8 ($\hat{r}=256$), this method is only 1.03 times slower than previous methods.

Inspired by RoPE~\cite{su2024roformer}, we can further refine this method by incorporating rotation operators into $f_{\text{comp}}$ to augment the expressiveness of $M$ by enable it to differentiate between various $x_{i\hat{r}: (i+1)\hat{r}}$ by rotating them.
We can modify Eq.~\ref{eq:decaq} as follows:
\begin{equation}
  \small
  \label{eq:rot}
\begin{split}
%a^i &= \text{rotx}\left(x_{i\hat{r}: (i+1)\hat{r}}, i \right)\\
%P^i &= \text{rotw}\left(M, i \right)\\
f_\text{comp}\left(x\right) &= \begin{bmatrix} a^1 & a^2 & \cdots  & a^{n-1} \end{bmatrix}\\
\Delta W &= \begin{bmatrix}
P^1 & \mathbf{0} & \cdots & \mathbf{0} \\
\mathbf{0} & P^2 & \cdots & \mathbf{0} \\
\vdots & \vdots & \ddots & \vdots \\
\mathbf{0} & \mathbf{0} & \cdots & P^{n-1}
\end{bmatrix}
\end{split}
\end{equation}
where $a^i$ and $P^i$ represent the corresponding values of $x_{i\hat{r}: (i+1)\hat{r}}$ and $M$ post-rotation, respectively.
Following RoPE, we use a $\hat{r}\times\hat{r}$ block diagonal matrix to achieve the rotation.
However, our method use rotation information to enable $M$ distinguish the $x_{i\hat{r}: (i+1)\hat{r}}$ instead of token position in RoPE.
We can define $a^i$ and $P^i$ as follows:
%\begin{equation}
%  \small
%  \label{eq:rot}
%\begin{split}
%  a^i &=  \begin{bmatrix}
%  \boldsymbol{R}_{\theta_1, i} & \mathbf{0} & \cdots & \mathbf{0} \\
%\mathbf{0} &   \boldsymbol{R}_{\theta_2, i} & \cdots & \mathbf{0} \\
%\vdots & \vdots & \ddots & \vdots \\
%\mathbf{0} & \mathbf{0} & \cdots & \boldsymbol{R}_{\theta_\frac{\hat{r}}{2}, i}
%\end{bmatrix} x_{i\hat{r}: (i+1)\hat{r}}\\
%  P^i &= M \begin{bmatrix}
%  \boldsymbol{R}_{\theta_1, i} & \mathbf{0} & \cdots & \mathbf{0} \\
%\mathbf{0} &   \boldsymbol{R}_{\theta_2, i} & \cdots & \mathbf{0} \\
%\vdots & \vdots & \ddots & \vdots \\
%\mathbf{0} & \mathbf{0} & \cdots & \boldsymbol{R}_{\theta_\frac{\hat{r}}{2}, i}
%\end{bmatrix}
%\end{split}
%\end{equation}
\begin{equation}
  \small
\begin{split}
  a^i &=  \begin{bmatrix}
  {R}_{\theta_1, i} & \mathbf{0} & \cdots & \mathbf{0} \\
\mathbf{0} &   {R}_{\theta_2, i} & \cdots & \mathbf{0} \\
\vdots & \vdots & \ddots & \vdots \\
\mathbf{0} & \mathbf{0} & \cdots & {R}_{\theta_\frac{\hat{r}}{2}, i}
\end{bmatrix} x_{i\hat{r}: (i+1)\hat{r}}\\
  P^i &= M \begin{bmatrix}
  {R}_{\theta_1, i} & \mathbf{0} & \cdots & \mathbf{0} \\
\mathbf{0} &   {R}_{\theta_2, i} & \cdots & \mathbf{0} \\
\vdots & \vdots & \ddots & \vdots \\
\mathbf{0} & \mathbf{0} & \cdots & {R}_{\theta_\frac{\hat{r}}{2}, i}
\end{bmatrix}
\end{split}
\end{equation}
where $\theta_j=10000^{-2(j-1) / \hat{r}}$ and ${R}_{\theta_j,i} \in \mathbb{R}^{2\times2}$ is a rotation matrix:
\begin{equation}
 {R}_{\theta_j, i}=\begin{bmatrix}
\cos i \theta_j & -\sin i \theta_j \\
\sin i \theta_j & \cos i \theta_j
\end{bmatrix}\\
\end{equation}

% \begin{bmatrix}
% x_1 \\
% x_2 \\
% x_3 \\
% x_4 \\
% \vdots \\
% x_{d-1} \\
% x_d
% \end{bmatrix} \otimes \begin{bmatrix}
% \cos m \theta_1 \\
% \cos m \theta_1 \\
% \cos m \theta_2 \\
% \cos m \theta_2 \\
% \vdots \\
% \cos m \theta_{d / 2} \\
% \cos m \theta_{d / 2}
% \end{bmatrix} + \begin{bmatrix}
% -x_2 \\
% x_1 \\
% -x_4 \\
% x_3 \\
% \vdots \\
% -x_d \\
% x_{d-1}
% \end{bmatrix} \otimes \begin{bmatrix}
% \sin m \theta_1 \\
% \sin m \theta_1 \\
% \sin m \theta_2 \\
% \sin m \theta_2 \\
% \vdots \\
% \sin m \theta_{d / 2} \\
% \sin m \theta_{d / 2}
% \end{bmatrix}

%\subsection{Merge able}
% \subsection{allow ranking increasing}

\section{Experiment}
We evaluate our method on various tasks to understand the influence of high-rank updating.
In Section~\ref{sec:memorizing}, we evaluate our method with LoRA and our method on memorizing UUID pairs to show the benefit of high-rank updating on memorizing.
In Section~\ref{sec:finetuning}, we reproduce LoRA, LoRA variants and FFT on three fine-tuning tasks: instruction tuning, mathematical reasoning and continual pretraining.
In Section~\ref{sec:pretraining}, we compare our method with LoRA and ReLoRA on pretraining by training transformer from scratch.
% Firstly, we compare our method on memorizing UUID key value pairs following Section~\ref{sec:3}, to check the improvement of memorizing bring by high-rank updating.
% Secondly, we evaluate three kinds of fine-tuning tasks: instruction tuning, mathematical reasoning and continual pretraining.
% Furthermore, we reproduce popular LoRA-varients such as DoRA, LoRA+ or AsymmetryLoRA on these tasks with 8 and 256 ranks, respectively.
%  we also discuss how trainable parameters influence the performance on three tasks and analyze the gap between FFT and LoRA.
%  Finally, we report the results on pretraining with LoRA and ReLoRA on 250M and 1B models from scratch. To check the influence of high-rank updating on pretraining.
\begin{figure}[t]
    \centering
        \includegraphics[width=0.99\columnwidth]{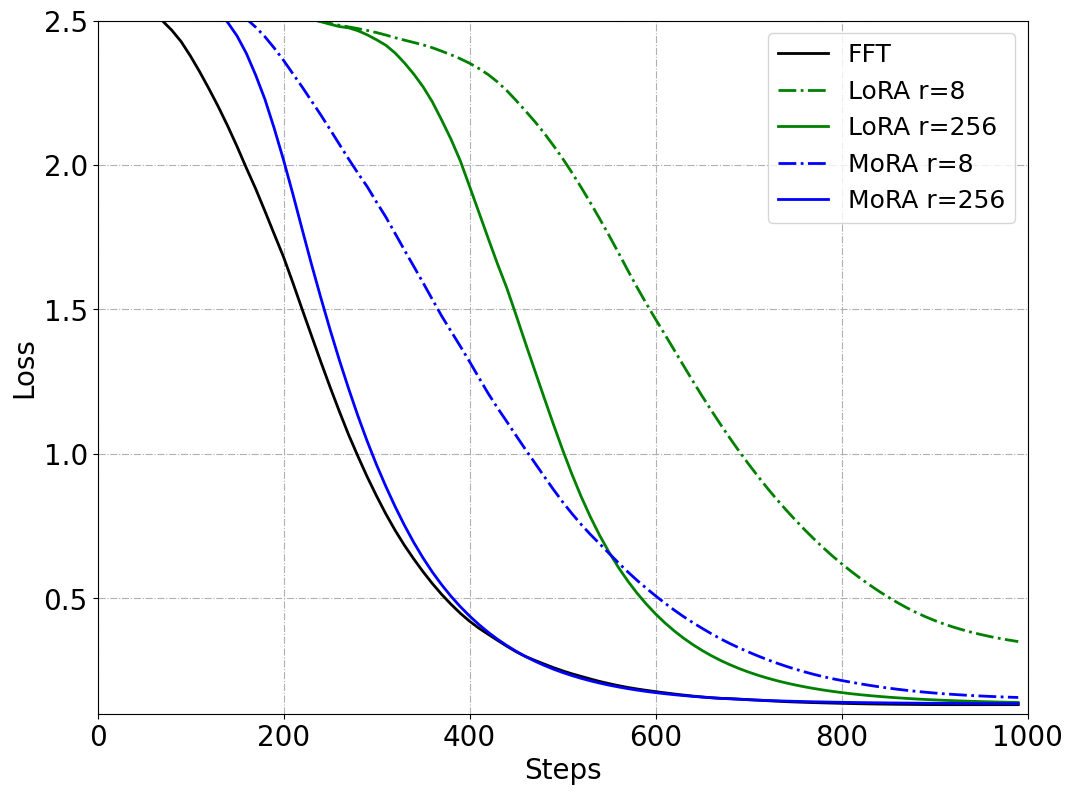}
    \caption{
      Performance of memorizing UUID pairs with LoRA and our method on rank 8 and 256.
      }
    \label{fig:moramtloss}
\vspace{-10pt}
\end{figure}

\begin{table*}[t]
\label{tab:main_results}
\centering
\begin{tabular}{lccccccc}
\toprule
&   & \multicolumn{2}{c}{Instruction Tuning} &   \multicolumn{2}{c}{Mathematical Reasoning} & \multicolumn{2}{c}{Continual Pretraining} \\ \cmidrule(lr){3-8}
%\textbf{Method} &  \textbf{\#Params\(\%\)} & \textbf{MMLU 0-shot} & \textbf{MMLU 5-shot} & \textbf{GSM8K} & \textbf{Math} & \textbf{Bio Avg.} \\
\textbf{Method} &  \textbf{Rank} & \textbf{MMLU 0} & \textbf{MMLU 5} & \textbf{GSM8K} & \textbf{MATH} & \textbf{BioMed.} &\textbf{Finance} \\
 \midrule
 FFT      & - & 50.6 & 51.3 & 66.6 & 20.1 & 56.4 & 69.6 \\
 \hline
 LoRA        &  8  & 50.2          &51.5          & \textbf{64.6}&15.1 &          52.3 & 64.0 \\
LoRA+        &  8  & 49.2          &51.1          & 64.1&\textbf{15.8} &          52.2 & 64.9 \\
ReLoRA        &  8  & 49.3          &50.2          & 61.5&14.5 &          46.3  & 61.0\\
AsyLoRA      &  8  & \textbf{50.3} &\textbf{52.2} & 64.5&15.0 &          52.5  & 63.5\\
DoRA         &  8  & 50.2          &51.5          & 64.5&14.6 &          52.5  & 63.9\\
MoRA (Ours)  &  8  & 49.7          &51.5          & 64.2&15.4 &          \textbf{53.3} & \textbf{67.1} \\
\hline
 LoRA        & 256 & 49.7          &50.8          & 67.9&\textbf{19.9} &          54.1 & 67.3 \\
 LoRA+       & 256 & 49.2          &51.3          & \textbf{68.2}&17.1 &          54.2 & 66.7\\
 ReLoRA       & 256 & -             & -            & 64.0&18.1 &          52.9  & 57.9\\
 AsyLoRA     & 256 & \textbf{50.1} &\textbf{52.0} & 66.9&19.3 &          54.1  & 66.9\\
 DoRA        & 256 & 49.6          &51.1          & 67.4&19.5 &          54.2 & 66.0 \\
 MoRA (Ours) & 256 & 49.9          &51.4          & 67.9&19.2 &          \textbf{55.4}  & \textbf{68.7}\\
\bottomrule
\end{tabular}
\caption{Performance of FFT, LoRA, LoRA variants and our method on instruction tuning, mathematical reasoning and continual pretraining tasks.}
\label{tab:main_results}
\end{table*}

\subsection{Memorizing UUID Pairs}
\label{sec:memorizing}
We first compare our method with LoRA and FFT on memorizing UUID pairs to demonstrate improvements through high-rank updating.
Following the training settings in Section~\ref{sec:3}, we search learning rate from $\{$5e-5$,$1e-4$,$2e-4$\}$ and use decompress and compress functions in Eq.~\ref{eq:decaq}, sharing rows and columns in $M$.
Due to use one matrix $M$ instead of two matrices $A$ and $B$, we can directly initialize $M$ with zeros.
For the predefined groups $g$ and $g'$, we group every adjacent $\hat{r}$ rows or columns together.
The training loss is presented in Figure\ref{fig:moramtloss}. Our method shows significant improvements over LoRA with the same number of trainable parameters, benefiting from high-rank updating. We also report character-level accuracy at various training steps in Table~\ref{your-table-label}. MoRA requires fewer training steps to memorize these UUID pairs compared to LoRA.
Compared to FFT, MoRA with 256 rank can achieve similar performance and both method can memorize all UUID pairs in 500 steps.

\begin{table}[t]
  \small
\centering
\begin{tabular}{lccccc}
\toprule
& Rank & {300} & {500} & {700} & {900} \\ \midrule
FFT & - & 42.5 & \cellcolor{lightcyan} 100 & \cellcolor{lightcyan} 100 & \cellcolor{lightcyan} 100 \\
LoRA & 8  & 9.9 & 10.0 & 10.7 & 54.2 \\
MoRA & 8   & 10.1 & 15.7 & 87.4 & \cellcolor{lightcyan}100 \\
LoRA & 256 & 9.9 & 70.6 & \cellcolor{lightcyan}100 & \cellcolor{lightcyan}100 \\
MoRA & 256 & 41.6 & \cellcolor{lightcyan}100 & \cellcolor{lightcyan}100 & \cellcolor{lightcyan}100 \\
\bottomrule
\end{tabular}
\caption{Character-level accuracy of memorizing UUID pairs by generating the value of corresponding key in 300, 500, 700 and 900 training steps.}
\label{your-table-label}
\vspace{-10pt}
\end{table}

\subsection{Fine-tuning Tasks}
\label{sec:finetuning}
\subsubsection{Setup}
We evaluate our method across three fine-tuning tasks for large language models (LLMs): instruction tuning, mathematical reasoning, and continual pretraining.
For these tasks, we select high-quality corresponding datasets to test both LoRA and our method.
In instruction tuning, we utilize Tülu v2~\cite{ivison2023camels}, a blend of several high-quality instruction datasets, containing 326k filtered samples. We assess instruction performance using the MMLU~\cite{hendrycks2020measuring} in both zero-shot and five-shot settings. For mathematical reasoning, we employ the MetaMath~\cite{yu2023metamath} with its 395k samples to enhance mathematical reasoning capabilities and also use GSM8K~\cite{cobbe2021training} and MATH~\cite{hendrycks2021measuring} for further evaluation. In continual pretraining, we adapt an LLM to the biomedicine and finance using PubMed abstracts from the Pile~\cite{gao2020pile} and finicial news, complemented by data preprocessing methods from AdaptLLM~\cite{cheng2023adapting} to boost performance. We report the average performance of corresponding tasks for continual pretraining. More details can be found in Appendix~\ref{sec:domain_adaptation}.
%The fine-tuning settings for these three tasks are detailed in Table~\ref{fine-tune-settings}.

% \begin{table}[h]
% \centering
% \small
% \begin{tabular}{lccccc}
% \toprule
%  & bsz & len & steps & epochs & tokens \\
% \midrule
% Tülu v2 & 128 & 2048 & 5k & 2 & 0.61B \\
% MetaMath & 128 & 512 & 9k & 3 & 109.5M \\
% PubMed & 128 & 2048 & 5k & 1 & 1.22B \\
% \bottomrule
% \end{tabular}
% \caption{Fine-tuning settings for three datasets with batch size, maximum input length, number of training steps, number of epochs, and total number of tokens in each dataset.}
% \label{table:fine-tune-settings}
% \end{table}

%\setlength{\parindent}{0pt}
%\textbf{Baselines and Implements}
%
\begin{figure*}[t]
    \centering
    \subfigure[Pretraining loss at 250M models.]
    {
        \includegraphics[width=1\columnwidth]{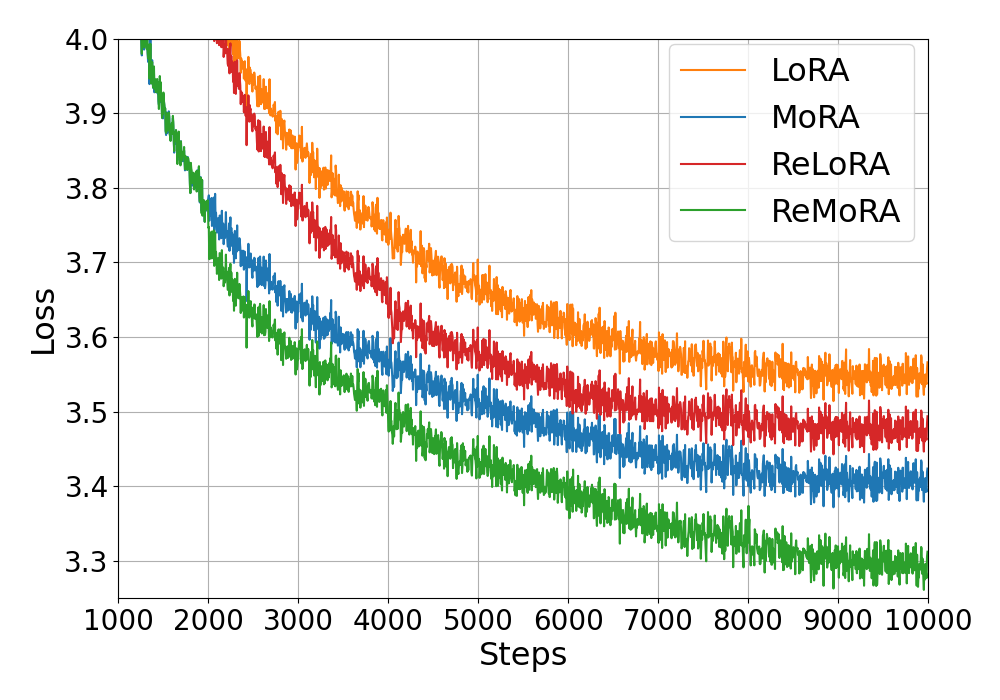}
    }
    \subfigure[Pretraining loss at 1.3B models.]
    {
        \includegraphics[width=1\columnwidth]{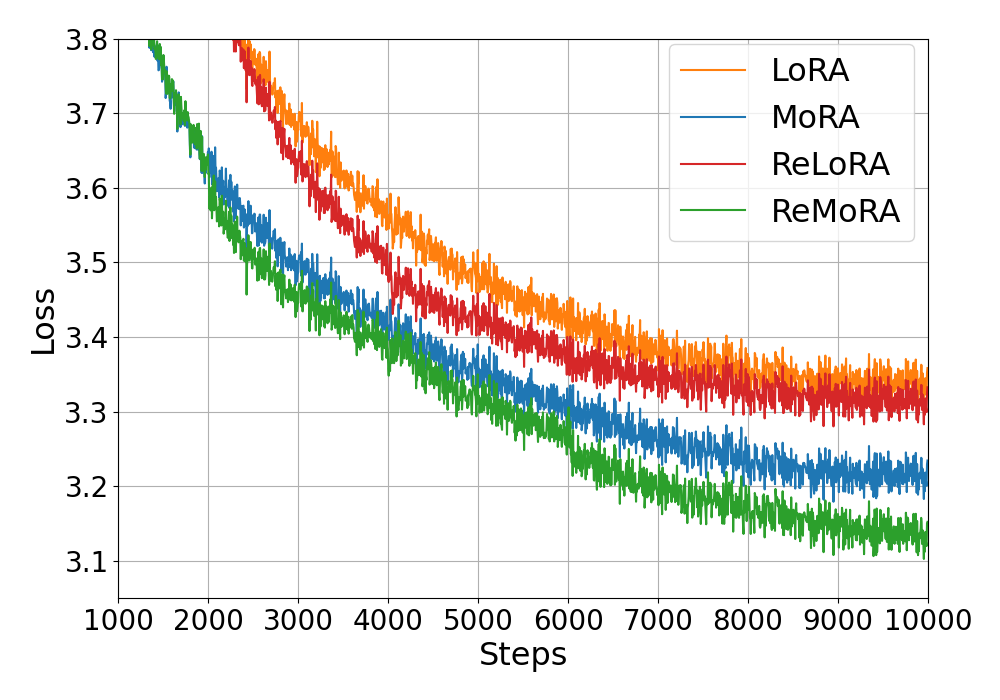}
    }
    \caption{
      Pretraining loss with LoRA and MoRA on 250M and 1B models from scratch.
      Both LoRA and MoRA use same amount of trainable parameters with $r=128$.
      ReMoRA and ReLoRA refer to merge MoRA or LoRA back to the model during training to increase the rank of $\Delta W$.
      }
    \label{fig:pretrain_loss}
\end{figure*}

\subsubsection{Baselines and Implements}
For LoRA-like methods and MoRA, we conducted experiments at $r=8$ and $r=256$, and reproduce following methods across three tasks: FFT, LoRA, LoRA+~\cite{Hayou2024}, AsyLoRA~\cite{zhu2024asymmetry}, ReLoRA~\cite{lialin2023stack} and DoRA~\cite{liu2024dora}.
LoRA+ enhances the learning rate of matrix $B$ in LoRA to facilitate efficient feature learning based on theoretical analysis.
We search the corresponding the hyperparameter $\lambda$ from $\{$2$, $4$\}$.
AsyLoRA also analyzes asymmetry in the $A$ and $B$ matrices, and we adopted their initialization strategy.
ReLoRA proposes a method to merge low-rank matrices into the model during training to increase the rank of $\Delta W$.
we search merge steps from \{$1$k, $2$k\} and use 50 steps restarts warmup.
DoRA leverages weight decomposition to enhance performance as a robust baseline.
For FFT, we follow the settings proposed by corresponding datasets.
For MoRA, we employed rotation operators as outlined in Eq.~\ref{eq:rot} to implement compression and decompression for $r=8$, and for $r=256$, we utilized shared rows and columns as specified in Eq.~\ref{eq:share} and group every adjacent $\hat{r}$ rows or columns together.
The details hyperparameters about fine-tuning can be found in Appendix~\ref{sec:train_details}.

%\setlength{\parindent}{0pt}
%\textbf{Results}
\subsubsection{Results and Analysis}
We present the results of fine-tuning tasks in Table~\ref{tab:main_results}.
We report the results of MMLU with zero-shot and 5-shot settings for instruction tuning, GSM8K and MATH for mathematical reasoning, and average performance on biomedical tasks and financial tasks for continual pretraining.

%\setlength{\parindent}{0pt}
%\textbf{MoRA show on par performance with LoRA.}
MoRA shows on par performances with LoRA on instruction tuning and mathematical reasoning.
Benefit from high-rank updating to memorize new knowledge, MoRA outperforms LoRA on both biomedical and financial domains for continual pretraining.
%Compared to LoRA, MoRA achieves high-rank updating with non-parameterized operators to comress and decompress features. where these operators

We also find that LoRA variants exhibit similar performances on these fine-tuning tasks as compared to LoRA. Although AsyLoRA achieves the best performance in instruction tuning, it demonstrates poor performance in mathematical reasoning. For ReLoRA, merging low-rank matrices during training can harm performance, particularly at the the high rank like 256.

Consider the difference between three tasks, they show different requirements for fine-tuning capabilities.
For instruction tuning, which does not learn new knowledge from fine-tuning, rank 8 is enough to achieve performance similar to FFT. For mathematical reasoning, rank 8 is unable to match FFT performance.
However, increasing the rank from 8 to 256 can eliminate the performance gap. For continual pretraining, LoRA with rank 256 still underperforms FFT.

\subsection{Pretraining}
\label{sec:pretraining}

\begin{table}[t]
%  \small
\centering
\begin{tabular}{lcc}
\toprule
 & 250M & 1.3B \\
\midrule
LoRA & 33.40 & 28.56 \\
MoRA (Ours) & 28.54 & 25.25 \\
ReLoRA & 32.19 & 27.80 \\
ReMoRA (Ours) & 26.74 & 23.34 \\
\bottomrule
\end{tabular}
\caption{Perplexity on C4 validation dataset.
   %Methods starting with "Re" utilize merging techniques during training.
}
\vspace{-10pt}
\label{table:ppl}
\end{table}

To understand the influence of high-rank updating,
we train transformer from scratch on the C4 datasets~\cite{raffel2020exploring}.
For the model architeture, we train LLaMA-based model with RMSNorm~\cite{zhang2019root}, SwiGLU~\cite{shazeer2020glu} and RoPE~\cite{su2024roformer} on 250M and 1.3B sizes.
For the hyperparameters, we use 10k steps, 1024 batch size, 512 sequence length and follow~\citeauthor{lialin2023stack}, using rank $r=128$ for LoRA and our methods and also keep modules without applying LoRA-like layernorm or embeddings unfreezed.
We compare our method with LoRA and ReLoRA.
To better show the difference between high-rank and low-rank updating, we reproduce ReLoRA and other methods without full-rank training warmup.
For MoRA, we use compression and decompression functions in Eq.~\ref{eq:share} by sharing columns and rows.

We also combine merge-and-reint in ReLoRA with our method called ReMoRA by merging $M$ back into the original parameters during training to increase the rank of $\Delta W$.
However, if we directly merge $M$ with $g$ and $g'$ in Eq.~\ref{eq:share}, the final rank of $\Delta W$ is unchanged due to the same expand pattern.
To solve this problem, we can change $g$ and $g'$ after merging to ensure the rank of $\Delta W$ increasing.
More details about ReMoRA can be found in Appendix~\ref{sec:ReMoRA}.
For the hyperparameters corresponding to ReLoRA and ReMoRA, we merge every 2k steps and use 50 steps restarts warmup with optimizer reseting and jagged scheduler.
% In this case, we simply define two kinds of $g$: $g^a$ and $g^b$.
% two kinds of $g$: \( g^a \) and \( g^b \). \( g^a \) groups every adjacent \( \hat{r} \) rows or columns together, the $i$-th groups is \(\{i, i+1, \ldots, i+\hat{r}\}\) like . Conversely, \( g^b \) groups elements starting from i and increases by \( r \) at each step, thereby the $i$-th groups like \(\{i, i+\hat{r}, \ldots, i+\hat{r}\hat{r}\}\).

We show pretraining loss in Figure~\ref{fig:pretrain_loss} and corresponding perplexity on C4 validation dataset in Table~\ref{table:ppl}.
Our method show better performance on pretraining compared to LoRA and ReLoRA with same amount of trainable parameters.
Benefiting from high-rank updating, ReMoRA also achieves more improvements on MoRA compared to ReLoRA, which
demonstrates the effectiveness of merge-and-reint strategy in ReMoRA.
%For the settings without merge-and-reint, MoRA achieve lower perplexity compared to LoRA.

\section{Analysis}

\begin{figure}[t]
\centering
    \includegraphics[width=1\columnwidth]{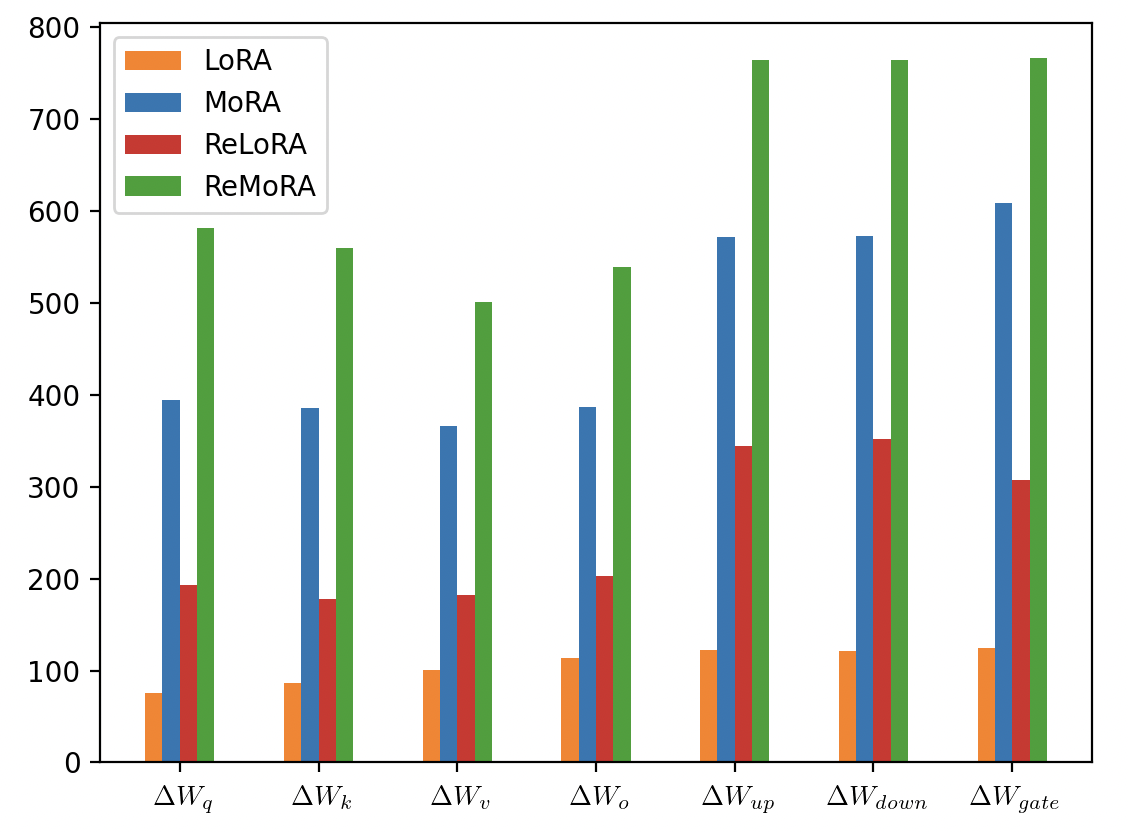}
\caption{
  The number of singular values $\textgreater 0.1$ in $\Delta W$ on the 250M pretraining model.
  }
 \label{fig:svd}
%\vspace{-10pt}
\end{figure}

\subsection{High-rank Updating}
To demonstrate the impact of high-rank updating on the rank of $\Delta W$, we analyzed the spectrum of singular values for the learned $\Delta W$ on 250M pretraining 250M model.
We present the average count of singular values exceeding 0.1 across all layers for $\Delta W_q$, $\Delta W_k$, $\Delta W_v$, $\Delta W_o$, $\Delta W_{up}$, $\Delta W_{down}$, and $\Delta W_{gate}$ in Figure~\ref{fig:svd} following~\cite{lialin2023stack}.
MoRA and ReMoRA exhibit a substantially higher number of significant singular values compared to LoRA and ReLoRA, highlighting the effectiveness of our methods in increasing the rank of $\Delta W$. We find the quantity of singular values shown in Figure~\ref{fig:svd} can be correlated with the perplexity metrics listed in Table~\ref{table:ppl}. Moreover, MoRA, without merge-and-reint strategy in ReLoRA and ReMoRA, can achieve a lower perplexity than ReLoRA along with a higher significant singular values.

\subsection{Influence of Decompression and Compression}

To explore the impact of decompression and compression functions in MoRA, we report the performance on GSM8K using various methods: truncation, sharing, decoupling, and rotation in Table~\ref{table:fun}.
Among these methods, truncation shows the worst performance due to the significant information loss during compression.
Sharing can achieve better performance than truncation by leveraging the shared rows or columns to preserve the input information.
But in the case of $r=8$, sharing shows worse performance than decouple and rotation due to the large number of sharing rows or columns, as we discussed in Section~\ref{sec:41}.
Rotation is more efficient than decouple, due to the rotation information can help the square matrix to distinguish the input information.

\begin{table}[h]
  \small
\centering
\begin{tabular}{cccc}
\toprule
& $f_{comp}$, $f_{decomp}$  & $r=8$ & $r=256$ \\
\midrule
Truncation & Eq.~\ref{eq:cut} & 59.5 &  66.6 \\
Sharing & Eq.~\ref{eq:share} & 62.5  &  67.9 \\
Decouple & Eq.~\ref{eq:decaq} & 63.6 & 67.8 \\
Rotation & Eq.~\ref{eq:rot} & 64.2   &  67.9\\
\bottomrule
\end{tabular}
\caption{Influence of decompression and compression functions on $r=8$ and $r=256$ on GSM8K.}
\label{table:fun}
\end{table}

\section{Conclusion}

In this paper, we analyze the impact of low-rank updating through LoRA and observe that such updating struggles for memory-intensive tasks, which also  limits current LoRA variants. To overcome this limitation, we introduce MoRA, a method that utilizes non-parameterized operators for high-rank updating. Within the MoRA framework, we explore various methods to implement decompression and compression functions. Performance comparisons indicate that MoRA matches LoRA in instruction tuning and mathematical reasoning, and exhibits superior performance in continual pretraining and memory tasks. Additionally, we conduct pretraining experiments to further demonstrate the effectiveness of high-rank updating and show superior results compared to ReLoRA.

\bibliography{custom}

\begin{thebibliography}{41}
\providecommand{\natexlab}[1]{#1}

\bibitem[{Chen et~al.(2023)Chen, Wang, Long, Zhang, Lu, Li, Wang, Xu, Bai, Huang et~al.}]{chen2023disc}
Wei Chen, Qiushi Wang, Zefei Long, Xianyin Zhang, Zhongtian Lu, Bingxuan Li, Siyuan Wang, Jiarong Xu, Xiang Bai, Xuanjing Huang, et~al. 2023.
\newblock Disc-finllm: A chinese financial large language model based on multiple experts fine-tuning.
\newblock \emph{arXiv preprint arXiv:2310.15205}.

\bibitem[{Chen et~al.(2022)Chen, Li, Smiley, Ma, Shah, and Wang}]{chen2022convfinqa}
Zhiyu Chen, Shiyang Li, Charese Smiley, Zhiqiang Ma, Sameena Shah, and William~Yang Wang. 2022.
\newblock Convfinqa: Exploring the chain of numerical reasoning in conversational finance question answering.
\newblock \emph{arXiv preprint arXiv:2210.03849}.

\bibitem[{Cheng et~al.(2023)Cheng, Huang, and Wei}]{cheng2023adapting}
Daixuan Cheng, Shaohan Huang, and Furu Wei. 2023.
\newblock Adapting large language models via reading comprehension.
\newblock \emph{arXiv preprint arXiv:2309.09530}.

\bibitem[{Cobbe et~al.(2021)Cobbe, Kosaraju, Bavarian, Chen, Jun, Kaiser, Plappert, Tworek, Hilton, Nakano et~al.}]{cobbe2021training}
Karl Cobbe, Vineet Kosaraju, Mohammad Bavarian, Mark Chen, Heewoo Jun, Lukasz Kaiser, Matthias Plappert, Jerry Tworek, Jacob Hilton, Reiichiro Nakano, et~al. 2021.
\newblock Training verifiers to solve math word problems.
\newblock \emph{arXiv preprint arXiv:2110.14168}.

\bibitem[{Collins et~al.(2023)Collins, Jiang, Frieder, Wong, Zilka, Bhatt, Lukasiewicz, Wu, Tenenbaum, Hart et~al.}]{collins2023evaluating}
Katherine~M Collins, Albert~Q Jiang, Simon Frieder, Lionel Wong, Miri Zilka, Umang Bhatt, Thomas Lukasiewicz, Yuhuai Wu, Joshua~B Tenenbaum, William Hart, et~al. 2023.
\newblock Evaluating language models for mathematics through interactions.
\newblock \emph{arXiv preprint arXiv:2306.01694}.

\bibitem[{Dernoncourt and Lee(2017)}]{dernoncourt2017pubmed}
Franck Dernoncourt and Ji~Young Lee. 2017.
\newblock Pubmed 200k rct: a dataset for sequential sentence classification in medical abstracts.
\newblock \emph{arXiv preprint arXiv:1710.06071}.

\bibitem[{Dettmers et~al.(2024)Dettmers, Pagnoni, Holtzman, and Zettlemoyer}]{dettmers2024qlora}
Tim Dettmers, Artidoro Pagnoni, Ari Holtzman, and Luke Zettlemoyer. 2024.
\newblock Qlora: Efficient finetuning of quantized llms.
\newblock \emph{Advances in Neural Information Processing Systems}, 36.

\bibitem[{Fu et~al.(2023)Fu, Peng, Ou, Sabharwal, and Khot}]{fu2023specializing}
Yao Fu, Hao Peng, Litu Ou, Ashish Sabharwal, and Tushar Khot. 2023.
\newblock Specializing smaller language models towards multi-step reasoning.
\newblock In \emph{International Conference on Machine Learning}, pages 10421--10430. PMLR.

\bibitem[{Gao et~al.(2020)Gao, Biderman, Black, Golding, Hoppe, Foster, Phang, He, Thite, Nabeshima et~al.}]{gao2020pile}
Leo Gao, Stella Biderman, Sid Black, Laurence Golding, Travis Hoppe, Charles Foster, Jason Phang, Horace He, Anish Thite, Noa Nabeshima, et~al. 2020.
\newblock The pile: An 800gb dataset of diverse text for language modeling.
\newblock \emph{arXiv preprint arXiv:2101.00027}.

\bibitem[{Han et~al.(2023)Han, Adams, Papaioannou, Grundmann, Oberhauser, L{\"o}ser, Truhn, and Bressem}]{han2023medalpaca}
Tianyu Han, Lisa~C Adams, Jens-Michalis Papaioannou, Paul Grundmann, Tom Oberhauser, Alexander L{\"o}ser, Daniel Truhn, and Keno~K Bressem. 2023.
\newblock Medalpaca--an open-source collection of medical conversational ai models and training data.
\newblock \emph{arXiv preprint arXiv:2304.08247}.

\bibitem[{Hayou et~al.(2024)Hayou, Ghosh, and Yu}]{Hayou2024}
Soufiane Hayou, Nikhil Ghosh, and Bin Yu. 2024.
\newblock \href {https://arxiv.org/abs/2402.12354} {{LoRA+: Efficient Low Rank Adaptation of Large Models}}.
\newblock 3.

\bibitem[{Hendrycks et~al.(2020)Hendrycks, Burns, Basart, Zou, Mazeika, Song, and Steinhardt}]{hendrycks2020measuring}
Dan Hendrycks, Collin Burns, Steven Basart, Andy Zou, Mantas Mazeika, Dawn Song, and Jacob Steinhardt. 2020.
\newblock Measuring massive multitask language understanding.
\newblock \emph{arXiv preprint arXiv:2009.03300}.

\bibitem[{Hendrycks et~al.(2021)Hendrycks, Burns, Kadavath, Arora, Basart, Tang, Song, and Steinhardt}]{hendrycks2021measuring}
Dan Hendrycks, Collin Burns, Saurav Kadavath, Akul Arora, Steven Basart, Eric Tang, Dawn Song, and Jacob Steinhardt. 2021.
\newblock Measuring mathematical problem solving with the math dataset.
\newblock \emph{arXiv preprint arXiv:2103.03874}.

\bibitem[{Houlsby et~al.(2019)Houlsby, Giurgiu, Jastrzebski, Morrone, De~Laroussilhe, Gesmundo, Attariyan, and Gelly}]{houlsby2019parameter}
Neil Houlsby, Andrei Giurgiu, Stanislaw Jastrzebski, Bruna Morrone, Quentin De~Laroussilhe, Andrea Gesmundo, Mona Attariyan, and Sylvain Gelly. 2019.
\newblock Parameter-efficient transfer learning for nlp.
\newblock In \emph{International conference on machine learning}, pages 2790--2799. PMLR.

\bibitem[{Hu et~al.(2021)Hu, Shen, Wallis, Allen-Zhu, Li, Wang, Wang, and Chen}]{hu2021lora}
Edward~J Hu, Yelong Shen, Phillip Wallis, Zeyuan Allen-Zhu, Yuanzhi Li, Shean Wang, Lu~Wang, and Weizhu Chen. 2021.
\newblock Lora: Low-rank adaptation of large language models.
\newblock \emph{arXiv preprint arXiv:2106.09685}.

\bibitem[{Imani et~al.(2023)Imani, Du, and Shrivastava}]{imani2023mathprompter}
Shima Imani, Liang Du, and Harsh Shrivastava. 2023.
\newblock Mathprompter: Mathematical reasoning using large language models.
\newblock \emph{arXiv preprint arXiv:2303.05398}.

\bibitem[{Ivison et~al.(2023)Ivison, Wang, Pyatkin, Lambert, Peters, Dasigi, Jang, Wadden, Smith, Beltagy et~al.}]{ivison2023camels}
Hamish Ivison, Yizhong Wang, Valentina Pyatkin, Nathan Lambert, Matthew Peters, Pradeep Dasigi, Joel Jang, David Wadden, Noah~A Smith, Iz~Beltagy, et~al. 2023.
\newblock Camels in a changing climate: Enhancing lm adaptation with tulu 2.
\newblock \emph{arXiv preprint arXiv:2311.10702}.

\bibitem[{Jin et~al.(2021)Jin, Pan, Oufattole, Weng, Fang, and Szolovits}]{jin2021disease}
Di~Jin, Eileen Pan, Nassim Oufattole, Wei-Hung Weng, Hanyi Fang, and Peter Szolovits. 2021.
\newblock What disease does this patient have? a large-scale open domain question answering dataset from medical exams.
\newblock \emph{Applied Sciences}, 11(14):6421.

\bibitem[{Jin et~al.(2019)Jin, Dhingra, Liu, Cohen, and Lu}]{jin2019pubmedqa}
Qiao Jin, Bhuwan Dhingra, Zhengping Liu, William Cohen, and Xinghua Lu. 2019.
\newblock Pubmedqa: A dataset for biomedical research question answering.
\newblock In \emph{Proceedings of the 2019 Conference on Empirical Methods in Natural Language Processing and the 9th International Joint Conference on Natural Language Processing (EMNLP-IJCNLP)}, pages 2567--2577.

\bibitem[{Kopiczko et~al.(2023)Kopiczko, Blankevoort, and Asano}]{kopiczko2023vera}
Dawid~Jan Kopiczko, Tijmen Blankevoort, and Yuki~Markus Asano. 2023.
\newblock Vera: Vector-based random matrix adaptation.
\newblock \emph{arXiv preprint arXiv:2310.11454}.

\bibitem[{Lester et~al.(2021)Lester, Al-Rfou, and Constant}]{lester2021power}
Brian Lester, Rami Al-Rfou, and Noah Constant. 2021.
\newblock The power of scale for parameter-efficient prompt tuning.
\newblock \emph{arXiv preprint arXiv:2104.08691}.

\bibitem[{Lialin et~al.(2023)Lialin, Shivagunde, Muckatira, and Rumshisky}]{lialin2023stack}
Vladislav Lialin, Namrata Shivagunde, Sherin Muckatira, and Anna Rumshisky. 2023.
\newblock Stack more layers differently: High-rank training through low-rank updates.
\newblock \emph{arXiv preprint arXiv:2307.05695}.

\bibitem[{Liu et~al.(2023)Liu, Ene, Kirby, Cheng, Pinckney, Liang, Alben, Anand, Banerjee, Bayraktaroglu et~al.}]{liu2023chipnemo}
Mingjie Liu, Teodor-Dumitru Ene, Robert Kirby, Chris Cheng, Nathaniel Pinckney, Rongjian Liang, Jonah Alben, Himyanshu Anand, Sanmitra Banerjee, Ismet Bayraktaroglu, et~al. 2023.
\newblock Chipnemo: Domain-adapted llms for chip design.
\newblock \emph{arXiv preprint arXiv:2311.00176}.

\bibitem[{Liu et~al.(2024)Liu, Wang, Yin, Molchanov, Wang, Cheng, and Chen}]{liu2024dora}
Shih-Yang Liu, Chien-Yi Wang, Hongxu Yin, Pavlo Molchanov, Yu-Chiang~Frank Wang, Kwang-Ting Cheng, and Min-Hung Chen. 2024.
\newblock Dora: Weight-decomposed low-rank adaptation.
\newblock \emph{arXiv preprint arXiv:2402.09353}.

\bibitem[{Maia et~al.(2018)Maia, Handschuh, Freitas, Davis, McDermott, Zarrouk, and Balahur}]{maia201818}
Macedo Maia, Siegfried Handschuh, Andr{\'e} Freitas, Brian Davis, Ross McDermott, Manel Zarrouk, and Alexandra Balahur. 2018.
\newblock Www'18 open challenge: financial opinion mining and question answering.
\newblock In \emph{Companion proceedings of the the web conference 2018}, pages 1941--1942.

\bibitem[{Malo et~al.(2014)Malo, Sinha, Korhonen, Wallenius, and Takala}]{malo2014good}
Pekka Malo, Ankur Sinha, Pekka Korhonen, Jyrki Wallenius, and Pyry Takala. 2014.
\newblock Good debt or bad debt: Detecting semantic orientations in economic texts.
\newblock \emph{Journal of the Association for Information Science and Technology}, 65(4):782--796.

\bibitem[{Meng et~al.(2024)Meng, Dai, Luo, Yang, Wu, Wang, Wang, Dong, Chen, and Sui}]{meng2024periodiclora}
Xiangdi Meng, Damai Dai, Weiyao Luo, Zhe Yang, Shaoxiang Wu, Xiaochen Wang, Peiyi Wang, Qingxiu Dong, Liang Chen, and Zhifang Sui. 2024.
\newblock Periodiclora: Breaking the low-rank bottleneck in lora optimization.
\newblock \emph{arXiv preprint arXiv:2402.16141}.

\bibitem[{Raffel et~al.(2020)Raffel, Shazeer, Roberts, Lee, Narang, Matena, Zhou, Li, and Liu}]{raffel2020exploring}
Colin Raffel, Noam Shazeer, Adam Roberts, Katherine Lee, Sharan Narang, Michael Matena, Yanqi Zhou, Wei Li, and Peter~J Liu. 2020.
\newblock Exploring the limits of transfer learning with a unified text-to-text transformer.
\newblock \emph{Journal of machine learning research}, 21(140):1--67.

\bibitem[{Renduchintala et~al.(2023)Renduchintala, Konuk, and Kuchaiev}]{renduchintala2023tied}
Adithya Renduchintala, Tugrul Konuk, and Oleksii Kuchaiev. 2023.
\newblock Tied-lora: Enhacing parameter efficiency of lora with weight tying.
\newblock \emph{arXiv preprint arXiv:2311.09578}.

\bibitem[{Salinas~Alvarado et~al.(2015)Salinas~Alvarado, Verspoor, and Baldwin}]{salinas-alvarado-etal-2015-domain}
Julio~Cesar Salinas~Alvarado, Karin Verspoor, and Timothy Baldwin. 2015.
\newblock \href {https://aclanthology.org/U15-1010} {Domain adaption of named entity recognition to support credit risk assessment}.
\newblock In \emph{Proceedings of the Australasian Language Technology Association Workshop 2015}, pages 84--90, Parramatta, Australia.

\bibitem[{Shazeer(2020)}]{shazeer2020glu}
Noam Shazeer. 2020.
\newblock Glu variants improve transformer.
\newblock \emph{arXiv preprint arXiv:2002.05202}.

\bibitem[{Shi et~al.(2024)Shi, Huang, Song, Li, Zhang, Huang, Wei, Deng, Sun, and Zhang}]{shi2024reslora}
Shuhua Shi, Shaohan Huang, Minghui Song, Zhoujun Li, Zihan Zhang, Haizhen Huang, Furu Wei, Weiwei Deng, Feng Sun, and Qi~Zhang. 2024.
\newblock Reslora: Identity residual mapping in low-rank adaption.
\newblock \emph{arXiv preprint arXiv:2402.18039}.

\bibitem[{Sinha and Khandait(2021)}]{sinha2021impact}
Ankur Sinha and Tanmay Khandait. 2021.
\newblock Impact of news on the commodity market: Dataset and results.
\newblock In \emph{Advances in Information and Communication: Proceedings of the 2021 Future of Information and Communication Conference (FICC), Volume 2}, pages 589--601. Springer.

\bibitem[{Su et~al.(2024)Su, Ahmed, Lu, Pan, Bo, and Liu}]{su2024roformer}
Jianlin Su, Murtadha Ahmed, Yu~Lu, Shengfeng Pan, Wen Bo, and Yunfeng Liu. 2024.
\newblock Roformer: Enhanced transformer with rotary position embedding.
\newblock \emph{Neurocomputing}, 568:127063.

\bibitem[{Wang et~al.(2018)Wang, Singh, Michael, Hill, Levy, and Bowman}]{wang2018glue}
Alex Wang, Amanpreet Singh, Julian Michael, Felix Hill, Omer Levy, and Samuel~R Bowman. 2018.
\newblock Glue: A multi-task benchmark and analysis platform for natural language understanding.
\newblock \emph{arXiv preprint arXiv:1804.07461}.

\bibitem[{Wang et~al.(2024)Wang, Ivison, Dasigi, Hessel, Khot, Chandu, Wadden, MacMillan, Smith, Beltagy et~al.}]{wang2024far}
Yizhong Wang, Hamish Ivison, Pradeep Dasigi, Jack Hessel, Tushar Khot, Khyathi Chandu, David Wadden, Kelsey MacMillan, Noah~A Smith, Iz~Beltagy, et~al. 2024.
\newblock How far can camels go? exploring the state of instruction tuning on open resources.
\newblock \emph{Advances in Neural Information Processing Systems}, 36.

\bibitem[{Wu et~al.(2023)Wu, Irsoy, Lu, Dabravolski, Dredze, Gehrmann, Kambadur, Rosenberg, and Mann}]{wu2023bloomberggpt}
Shijie Wu, Ozan Irsoy, Steven Lu, Vadim Dabravolski, Mark Dredze, Sebastian Gehrmann, Prabhanjan Kambadur, David Rosenberg, and Gideon Mann. 2023.
\newblock Bloomberggpt: A large language model for finance.
\newblock \emph{arXiv preprint arXiv:2303.17564}.

\bibitem[{Yu et~al.(2023)Yu, Jiang, Shi, Yu, Liu, Zhang, Kwok, Li, Weller, and Liu}]{yu2023metamath}
Longhui Yu, Weisen Jiang, Han Shi, Jincheng Yu, Zhengying Liu, Yu~Zhang, James~T Kwok, Zhenguo Li, Adrian Weller, and Weiyang Liu. 2023.
\newblock Metamath: Bootstrap your own mathematical questions for large language models.
\newblock \emph{arXiv preprint arXiv:2309.12284}.

\bibitem[{Zhang and Sennrich(2019)}]{zhang2019root}
Biao Zhang and Rico Sennrich. 2019.
\newblock Root mean square layer normalization.
\newblock \emph{Advances in Neural Information Processing Systems}, 32.

\bibitem[{Zhou et~al.(2024)Zhou, Liu, Xu, Iyer, Sun, Mao, Ma, Efrat, Yu, Yu et~al.}]{zhou2024lima}
Chunting Zhou, Pengfei Liu, Puxin Xu, Srinivasan Iyer, Jiao Sun, Yuning Mao, Xuezhe Ma, Avia Efrat, Ping Yu, Lili Yu, et~al. 2024.
\newblock Lima: Less is more for alignment.
\newblock \emph{Advances in Neural Information Processing Systems}, 36.

\bibitem[{Zhu et~al.(2024)Zhu, Greenewald, Nadjahi, Borde, Gabrielsson, Choshen, Ghassemi, Yurochkin, and Solomon}]{zhu2024asymmetry}
Jiacheng Zhu, Kristjan Greenewald, Kimia Nadjahi, Haitz S{\'a}ez de~Oc{\'a}riz Borde, Rickard~Br{\"u}el Gabrielsson, Leshem Choshen, Marzyeh Ghassemi, Mikhail Yurochkin, and Justin Solomon. 2024.
\newblock Asymmetry in low-rank adapters of foundation models.
\newblock \emph{arXiv preprint arXiv:2402.16842}.

\end{thebibliography}

\newpage
\onecolumn
\appendix

\section{Hyperparameters}
\label{sec:train_details}

We propose hyperparameters in Table~\ref{table:fine-tune-settings}.

% \begin{table*}[h]
% \centering
% \begin{tabular}{ccc}
% \toprule
% Hyperparameters  &\\
% \midrule
% Rank $r$ & 8  & \cellcolor{lightcyan}256 \\
% $\alpha$ & 16 & 128 \\
% Dropout & \multicolumn{2}{c}{0.05}\\
% Optimizer & \multicolumn{2}{c}{AdamW}\\
% LR & \multicolumn{2}{c}{$\{$1e-4$,$2e-4$\}$}\\
% LR  Scheduler & \multicolumn{2}{c}{cosine}\\
% Warmup Steps & \multicolumn{2}{c}{150} \\
% Epochs & \multicolumn{2}{c}{2} \\
% Batch size & \multicolumn{2}{c}{128} \\
% Where &\multicolumn{2}{c}{Q,K,V,O,Up,Down,Gate} \\
% \bottomrule
% \end{tabular}
% \caption{Fine-tuning settings for three datasets with batch size, maximum input length, number of training steps, number of epochs, and total number of tokens in each dataset.}
% \label{table:fine-tune-settings}
% \end{table*}

\begin{table*}[h]
  \small
\centering
\begin{tabular}{lccccccccc}
\toprule
Dataset                   & Method    & $r$ & $\alpha$ & LR                  & LR Scheduler & Warmup & Epochs & Batch size & $f_{\text{comp}}$, $f_{\text{decomp}}$  \\ \midrule
\multirow{5}{*}{Tülu v2}  & FFT       & -   & -   & 2e-5                & cosine       & 500    & 2      & 128        & -  \\
                          & LoRA-like & 8   & 16  & $\{$1e-4$,$2e-4$\}$ & cosine       & 500    & 2      & 128        & - \\
                          & MoRA      & 8   & -   & $\{$2e-4$,$3e-4$\}$ & cosine       & 500    & 2      & 128        & Eq.~\ref{eq:rot} \\
                          & LoRA-like & 256 & 128 & $\{$1e-4$,$2e-4$\}$ & cosine       & 500    & 2      & 128        & -\\
                          & MoRA      & 256 & -   & $\{$3e-5$,$5e-5$\}$ & cosine       & 500    & 2      & 128        & Eq.~\ref{eq:share} \\
\midrule
\multirow{5}{*}{MetaMath} & FFT       & -   & -   & 2e-5                & cosine       & 300    & 3      & 128        & -  \\
                          & LoRA-like & 8   & 16  & $\{$1e-4$,$2e-4$\}$ & cosine       & 300    & 3      & 128        & - \\
                          & MoRA      & 8   & -   & $\{$2e-4$,$3e-4$\}$ & cosine       & 300    & 3      & 128        & Eq.~\ref{eq:rot} \\
                          & LoRA-like & 256 & 128 & $\{$1e-4$,$2e-4$\}$ & cosine       & 300    & 3      & 128        & -\\
                          & MoRA      & 256 & -   & $\{$3e-5$,$5e-5$\}$ & cosine       & 300    & 3      & 128        & Eq.~\ref{eq:share} \\
\midrule
\multirow{5}{*}{BioMed./Fiance}   & FFT       & -   & -   & 3e-5                & linear       & 150    & 3      & 128        & -  \\
                          & LoRA-like & 8   & 16  & $\{$3e-4$,$4e-4$\}$ & linear       & 150    & 3      & 128        & - \\
                          & MoRA      & 8   & -   & $\{$4e-4$,$5e-4$\}$ & linear       & 150    & 3      & 128        & Eq.~\ref{eq:rot} \\
                          & LoRA-like & 256 & 128 & $\{$3e-4$,$4e-4$\}$ & linear       & 150    & 3      & 128        & -\\
                          & MoRA      & 256 & -   & $\{$5e-5$,$7e-5$\}$ & linear       & 150    & 3      & 128        & Eq.~\ref{eq:share} \\
\bottomrule
\end{tabular}
\caption{Hyperparameters for fine-tuning on three datasets.}
\label{table:fine-tune-settings}
\end{table*}
%For the $g$ and $g'$ in Eq.~\ref{eq:share}, we use $g_i=\{(i-1)\hat{r}+1,\dots,i\hat{r}\}$ and $g_i'=\{(i-1)\hat{r}+1,\dots,i\hat{r}\}$.

\section{Implementation of ReMoRA}
\label{sec:ReMoRA}
We introduce detial implementation of ReMoRA in pretraining.
In this case, we simply define two kinds of $g$.
The first kind is grouping every adjacent \( \hat{r} \) rows or columns together following the defined in fine-tuning, the first groups can be represented as \(\{1, 2, \ldots, \hat{r}\}\).
The second kind is grouping every neighboring $k$ of the rows or columns together, the first groups can be represented as \(\{1, 1+k, \ldots, 1+\hat{r}k\}\).
We propose a example code about compression and decompression functions in Algorithm~\textbf{1} and \textbf{2}.
After merging, we can change the group type from $0$ to $1$ or $1$ to $0$.

\begin{algorithm}
  \label{alg:compress}
  \small
\caption{Compression}
\begin{algorithmic}[1]
\Function{compress}{$x$, $\hat{r}$, $type$}
    \State \# $x\in\mathbb{R}^{bsz \times l  \times k}$: Input tensor
    \State \# $y\in\mathbb{R}^{bsz \times l  \times \hat{r}}$: Output tensor
    \State \# $type \in \{0, 1\}$: Group type 0 or 1
    \State padding $x$ to make $k$ divisible by $\hat{r}$
    \If {$type=0$}
          \State $y$ = $x$.view($bsz, l, k/\hat{r}, \hat{r}$).sum(dim=$2$) \# first type of group
    \Else
          \State $y$ = $x$.view($bsz, l, \hat{r}, k/\hat{r}$).sum(dim=$3$) \# second type of group
    \EndIf
    \State \textbf{return} $y$
\EndFunction
\end{algorithmic}
\end{algorithm}

\begin{algorithm}
  \label{alg:decompress}
  \small
\caption{Decompression}
\begin{algorithmic}[1]
\Function{decompress}{$x$, $\hat{r}$, $type$}
    \State \# $x\in\mathbb{R}^{bsz \times l  \times \hat{r}}$: Input tensor
    \State \# $y\in\mathbb{R}^{bsz \times l  \times d}$: Output tensor
    \State \# $type \in \{0, 1\}$: Group type 0 or 1
    \If {$type=0$}
          \State $y$ = repeat(x, $d/\hat{r}$, dim=$2$) \# first type of group
    \Else
          \State $y$ = repeat-interleave(x, $d/\hat{r}$, dim=$2$) \# second type of group
    \EndIf
    \State truncate $y$ to $\mathbb{R}^{bsz \times l  \times d}$
    \State \textbf{return} $y$
\EndFunction
\end{algorithmic}
\end{algorithm}

\section{Downstream Tasks of Continual Pretraining}
\label{sec:domain_adaptation}
For biomedcine, we use PubMedQA~\cite{jin2019pubmedqa}, RCT~\cite{dernoncourt2017pubmed}, USMLE~\cite{jin2021disease}, and selecting biomedicine subjects from MMLU to evaluate the performance.
For finance, following BloombergGPT~\cite{wu2023bloomberggpt},we use ConvFinQA~\cite{chen2022convfinqa}, NER~\cite{salinas-alvarado-etal-2015-domain}, Headline~\cite{sinha2021impact}, FiQA SA~\cite{maia201818} and FPB~\cite{malo2014good}.
We report the detail performance of these tasks following:

\begin{table}[h]
\label{table:bio}
\small
\centering
 \begin{tabularx}{0.9\textwidth}{XYYYYYY}
%\begin{tabular}{lcccccc}
  \toprule
    & $r$& \textbf{PubMedQA} & \textbf{USMLE} & \textbf{BioMMLU} & \textbf{RCT} & \textbf{Avg.} \\ \hline
FFT  & -   & 74.1 & 41.2 & 47.5 & 62.7 & 56.4 \\
LoRA & 8   & 73.1 & 34.9 & 45.3 & 54.9 & 51.9 \\
MoRA & 8   & 73.3 & 34.7 & 45.3 & 59.9 & 53.3 \\
LoRA & 256 & 73.8 & 39.7 & 46.0 & 56.9 & 54.1 \\
MoRA & 256 & 74.4 & 40.4 & 46.1 & 60.6 & 55.4 \\
\bottomrule
%\end{tabular}
\end{tabularx}
\caption{Performance on biomedical tasks.}
\end{table}

\begin{table}[h]
\label{table:fin}
\small
\centering
 \begin{tabularx}{0.90\textwidth}{XYYYYYYY}
  \toprule
& $r$    & \textbf{ConvFinQA} & \textbf{FiQA SA} & \textbf{Headline} & \textbf{NER} & \textbf{FPB}& \textbf{Avg.} \\ \hline
FFT  & -   & 44.4 & 78.8 & 82.3 & 68.1 & 74.3 & 69.6 \\
LoRA & 8   & 44.5 & 76.2 & 72.4 & 61.6 & 65.1 &64.0 \\
MoRA & 8   & 45.8 & 76.6 & 76.3 & 68.9 & 68.2 &67.1 \\
LoRA &  256& 41.4 & 78.3 & 83.0 & 66.8 & 66.7 & 67.3  \\
MoRA & 256 & 47.7 & 76.3 & 83.4 & 68.0 & 68.1 & 68.7 \\
\bottomrule
\end{tabularx}
\caption{Performance on finicial tasks.}
\end{table}

%\label{sec:appendix}

%This is an appendix.

\end{document}